\documentclass[sigconf]{acmart}
\usepackage{helvet} % DO NOT CHANGE THIS
\usepackage{courier}  % DO NOT CHANGE THIS
\usepackage{graphicx}  % DO NOT CHANGE THIS
\usepackage{booktabs}
\usepackage{pifont}
\usepackage{float}
\usepackage{epstopdf}
\usepackage{amssymb}
\usepackage{algorithm}
\usepackage{algorithmicx}
\usepackage[noend]{algpseudocode}
\usepackage{array}
\usepackage{amsthm,amssymb,amsfonts}
\usepackage{multirow}
\usepackage{subfig}
\usepackage{multicol}
\usepackage{fixltx2e}
\usepackage{stfloats}
\usepackage{dirtytalk}
\usepackage{mathrsfs}
\usepackage{bm}

\AtBeginDocument{%
  \providecommand\BibTeX{{%
    \normalfont B\kern-0.5em{\scshape i\kern-0.25em b}\kern-0.8em\TeX}}}

\setcopyright{acmcopyright}
\copyrightyear{2020}
\acmYear{2020}
\acmDOI{10.1145/3340531.3411913}

\acmConference[CIKM '20]{CIKM '20: ACM International Conference on Information and Knowledge management}{October 19--23, 2020}{Online}
\acmBooktitle{CIKM '20: ACM International Conference on Information and Knowledge management, October 19--23, 2020, Online}
\acmPrice{15.00}
\acmISBN{978-1-4503-6859-9/20/10}

\begin{document}
\fancyhead{}

\title{Auxiliary-task Based Deep Reinforcement Learning for Participant Selection Problem in Mobile Crowdsourcing}
\author{Wei Shen}
\authornote{These authors contributed equally to this research.}
\email{shenwei0917@126.com}
\affiliation{
  \institution{Baidu Inc.}
}

\author{Xiaonan He}
\authornotemark[1]
\email{xiaonan.cs@gmail.com}
\affiliation{
  \institution{Baidu Inc.}
}

\author{Chuheng Zhang}
\authornotemark[1]
\email{zhangchuheng123@live.com}
\affiliation{
  \institution{IIIS, Tsinghua University}
}

\author{Qiang Ni}
\email{q.ni@lancaster.ac.uk}
\affiliation{
  \institution{Lancaster University}
}

\author{Wanchun Dou}
\authornote{Corresponding author.}
\email{douwc@nju.edu.cn}
\affiliation{
  \institution{Nanjing University}
}

\author{Yan Wang}
\email{reggiewang@tencent.com}
\affiliation{
  \institution{Tencent Holdings Ltd.}
}

\renewcommand{\shortauthors}{Anonymous Author(s)}

\begin{abstract}
In mobile crowdsourcing (MCS), the platform selects participants to complete location-aware tasks from the recruiters aiming to achieve multiple goals (e.g., profit maximization, energy efficiency, and fairness). 
However, different MCS systems have different goals and there are possibly conflicting goals even in one MCS system. 
Therefore, it is crucial to design a participant selection algorithm that applies to different MCS systems to achieve multiple goals. 
To deal with this issue, we formulate the participant selection problem as a reinforcement learning problem and propose to solve it with a novel method, which we call auxiliary-task based deep reinforcement learning (ADRL). 
We use transformers to extract representations from the context of the MCS system and a pointer network to deal with the combinatorial optimization problem. 
To improve the sample efficiency, we adopt an auxiliary-task training process that trains the network to predict the imminent tasks from the recruiters, which facilitates the embedding learning of the deep learning model. 
Additionally, we release a simulated environment on a specific MCS task, the ride-sharing task, and conduct extensive performance evaluations in this environment.
The experimental results demonstrate that ADRL outperforms and improves sample efficiency over other well-recognized baselines in various settings.
\end{abstract}



\maketitle

\section{Introduction}
With the massive deployment of mobile devices, mobile crowdsourcing (MCS) has become a new service paradigm in recent years. 
The MCS platforms recruit a large number of participants to supply fast, economical and efficient services for mobile users, including online product searching \cite{Fan2016CrowdOp}, sports video categorization \cite{Agadakos2017Techu}, and ride-sharing \cite{Suhr:2019}. Furthermore, given the explosive growth of data available on the web, MCS can help users to obtain effective information in many online services such as online shopping sites. All in all, MCS systems have been playing a vital and indispensable role in various areas to boost business and are pervasive across numerous online services.

There are two main roles in a typical MCS system, recruiters and participants. The recruiters are those who outsource their tasks to the crowd and the participants are those who are willing to accomplish the tasks. The undertaking of the tasks brings the participants profit, knowledge, and experience \cite{Mao2016A}, and also meets the requirements of the recruiters. 
Accordingly, how to select suitable participants for each task is an important research topic that many researchers focus on \cite{Tong2016Online}.

On an MCS platform, the participant selection (or task allocation) method is important for the platform to improve the efficiency, increase the overall profit, and attract more recruiters and participants.
Nevertheless, conventional participant selection methods \cite{Tong2016Online,Pan2016Efficient} can hardly achieve a good performance in a multi-goal setting where different goals such as fairness, energy efficiency, and profit maximization need to be optimized simultaneously. 
In a ride-sharing system, a specific MCS system, shown in Figure \ref{fig:Motivation}, diverse objects ask for diverse participant selection methods. 
When the objective is to minimize the total time consumption (Figure \ref{fig:Motivation}a), the participant $V3$ is selected according to the distance factor which is a key factor for time consumption. 
When the objective is changed to fairness (Figure \ref{fig:Motivation}b), the participant $V4$ is chosen to complete a new task because it receives the least incentives in previous tasks. 
Under such contexts, the incentives a participant receives is a key factor that affects the participant selection strategy. 
When the objective is changed to minimize the total energy consumption (Figure \ref{fig:Motivation}c), $V1$ is selected because it consumes the lowest energy under such a circumstance. 
However, when the goal is a mixture of the above goals, (Figure \ref{fig:Motivation}d), it is hard to find a key factor or a simple participant selection method to optimize all of these objectives. 
Conventional participant selection methods become ineffective in this situation. 
Moreover, environment uncertainty such as traffic jams also brings difficulties for conventional participant selection methods. As such, in this paper, we address a key research issue in MCS: how to select participants to complete the tasks in a multi-goal environment? 

The method proposed by \citeauthor{Pu2016Crowdlet} \cite{Pu2016Crowdlet} requires to accumulate the personal information of participants to predict their mobile patterns. However, it is hard for the platform to obtain the full information of a participant. 
In addition, the trade-off between different goals is fixed and naive in previous researches. 
For instance, \citeauthor{peng2017fair} \cite{peng2017fair} present a trade-off between fairness and energy-efficiency, which is characterized by min-max aggregate sensing time. 
Compared to the above methods, reinforcement learning is more suitable for a multi-goal environment. 
\citeauthor{Sadhu2017Argus} \cite{Sadhu2017Argus} adopt deep reinforcement learning to maximize the expected pay-off against the uncertainty in MCS. 
\citeauthor{lin2018efficient} \cite{lin2018efficient} use multi-agent reinforcement learning to predict explicit coordination among participants. 
However, the multi-goal setting in MCS is not considered in these papers.
To the best of our knowledge, this is the first paper that proposes an end-to-end participant selection method via ADRL to solve the multi-goal optimization problem in MCS. There are three challenges in the participant selection problem with multi-goal setting:

\begin{figure*}[!t]
    \centering\includegraphics[width=6.5in]{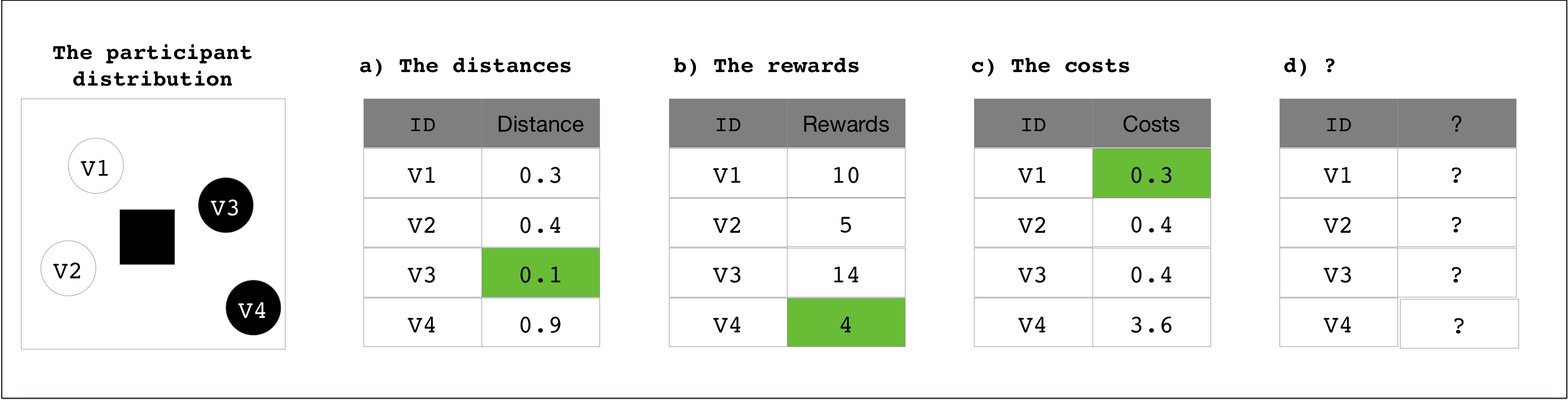}
    \caption{A motivation example in a specific MCS system, ride-sharing system. }
    \label{fig:Motivation}
\end{figure*}

\textbf{Conflicting goals.} 
The motivation examples disclose that, given a mixture of goals in MCS, it is hard to find a key factor or a simple participant selection method. Moreover, in the multi-goal setting, the multiple goals are possibly conflicting with each other. For instance, achieving the goal of fairness might lead to a decrease in the total profit and energy efficiency. It is a challenge to trade-off among the various goals.

\textbf{Structured context in MCS.} The context of a typical MCS environment is structured (e.g., the context is represented by the timestamps, the text data and the position information), which makes it difficult to distill information from the underlying intricate relationships. 

\iffalse
The goal varies greatly across different MCS systems. For instance, in an environmental monitor system, the participants equipped with mobile sensors upload the environmental data. One of the most important goals of the system is energy efficiency since the tasks consume the power of the equipment for sensing and monitoring. However, in a ride-sharing system, we pay more attention to profit maximization and fairness than energy efficiency. It is a challenge to design a unified algorithm for different systems with different goals.
\fi

%multi-type task

\textbf{Sample complexity in reinforcement learning.}
The model-free reinforcement learning algorithms usually suffer from sample inefficiency, especially when the policy is approximated by a neural network with a large number of parameters. Since the samples in real scenarios are expensive, it is hard to learn a policy by a sample inefficient algorithm. 
Meanwhile, the goal is flexible (i.e., the multiple goals or the trade-off between different goals of an MCS system may change over time) and therefore we need to frequently retrain the model. 
However, training a model using a sample inefficient algorithm is time-consuming even in a simulated environment.

In response to the challenges mentioned above, we propose a participant selection method using ADRL based on transformers and a pointer network, inspired by AlphaStar \cite{arulkumaran2019alphastar}. More specifically, we adopt the transformers to extract features from the participants and the tasks, and the pointer network to select suitable participants to complete the tasks. The main contributions of our work are summarized as follows:

\begin{itemize}
    \item We formulate the participant selection problem as a reinforcement learning problem which enables optimizations for arbitrary reward functions and thus is suitable for the flexible and multiple goal setting.
    \item We release a simulated MCS environment based on real datasets, which can generate a large number of experiences to train policies under the reinforcement learning formulation. 
    Besides, the configurations of the tasks, the participants and other environment conditions can be easily customized. 
    For the ease of future research, we base our simulated environment on the OpenAI Gym \cite{brockman2016openai} interface and publish it on \url{https://github.com/swtheing/Auxiliary-task-Based-Deep-Reinforcement-Learning}.
    \item In ADRL, we use the transformers to exploit the inherent structures in the MCS environment and use the pointer network to select suitable participants to complete the tasks.
    \item We design an auxiliary task training process to improve sample efficiency and therefore accelerate the training. 
    In this process, we train the model to predict the imminent tasks in the future, which helps the model to learn a better representation of tasks and participants. 
    Empirically, we observe that this process greatly improves the sample efficiency and the performance of the policy.
\end{itemize}

The rest of this paper is organized as follows. The problem statement and the simulated environment are introduced in Section 2. The policy network of ADRL is specified in Section 3. The auxiliary training process is shown in Section 4. Experiments based on the real dataset and performance evaluations are presented in Section 5. The related work and conclusion are shown in Section 6 and Section 7 respectively.

\section{Problem Statement and Simulated Environment}

In this section, we formulate the participant selection problem as a reinforcement learning problem and introduce the simulated environment for ride-sharing, which is a specific MCS task.

\subsection{Problem Statement}
In an MCS system, a recruiter requests a location-aware task at a time point. Then, the platform accumulates a batch of tasks in a time interval. At the end of the time interval, the platform assigns the tasks to suitable participants. Under such a context, the participant selection problem in MCS is to select suitable participants to complete these tasks for multiple goals. In this problem, the platform makes decisions (select suitable participants for the tasks) at the end of each time interval and the decision in a particular time interval affects the subsequent decisions.
This makes it hard to divide this problem into a series of subproblems, each of which corresponds to a particular time interval. 
Therefore, we formulate the participant selection problem as a Markov decision problem in this paper. 

The state and the action of Markov decision problem are defined as follows:
\begin{equation}
\begin{aligned}
s_h  = &\{\bm{FT}_h,\bm{FP}_h, \bm{FE}_h \} \\
a_h  = & \{a_{ht}\}_{t \in \bm{T}_h}\\
\end{aligned}
\end{equation}
where the subscript $h$ represents the index of the time interval.
The state $s_h$ consists of three parts, the task features $\bm{FT}_h$, the participant features $\bm{FP}_h$ and the environment features $\bm{FE}_h$ at the end of the $h$-th time interval.
The set of the pending tasks collected by the platform till the end of the $h$-th time interval is denoted as $\bm{T}_h$, and the corresponding set of available participants is denoted as $\bm{P}_h$.
The platform selects one or more participants for each task $t$ in $\bm{T}_h$ from the set of all currently available participants at the end of the $h$-th time interval, which is denoted as $a_{ht}\subset \bm{P}_h$. The set of the selections for all the tasks in the $h$-th time interval is defined as the action $a_h$.
Thus, the participant selection policy is defined as $\pi (a_h|s_h)$. 
At last, the reward for the Markov decision problem can be set arbitrarily to indicate the goals of the system.

In the experiments, we use the following features: 
The task features include basic task description, timestamp, recruiter information, location information, and the requirements for the task (e.g., time limit and the number of participants required).
The participant features include basic participant description, location information and the state of the participant (e.g., whether the participant is occupied).
The environment features include the weather condition and the traffic condition.
In our implementation, the environment features for the locations associated with participants or tasks are appended to the corresponding participant or task representations.

The reward of the sequential decision process is designed and serves as the objective for the policy optimization. We design a flexible reward function for the multi-goal setting,
\begin{equation}
    r_h(\cdot) = \sum_{h}\sum_{k}w_k o_{hk}(\cdot),
\end{equation}
where $r_h$ is the reward function in the $h$-th time interval,
$o_{hk}(\cdot)$ is function of the $k$-th objective in the $h$-th time interval and $w_k$ is the weight of the objective $o_{hk}$. 

There are three main goals in our setting specified as follows: 

\begin{itemize}
    \item \textbf{Effectiveness.} Participants consume resources to accomplish the tasks. Therefore, the platform needs to control the total amount of the resources to level up the effectiveness. 
    \item \textbf{Fairness.} The platform needs to allocate each participant roughly an equal number of tasks, which helps the participants to make profits and incentivizes the participants to take more tasks.
    \item \textbf{Overall profit.} The platform needs to make the overall profits to accelerate its development. 
\end{itemize}

\subsection{Simulated Environment}

\begin{figure}[ht]
    \centering\includegraphics[width=3.2in]{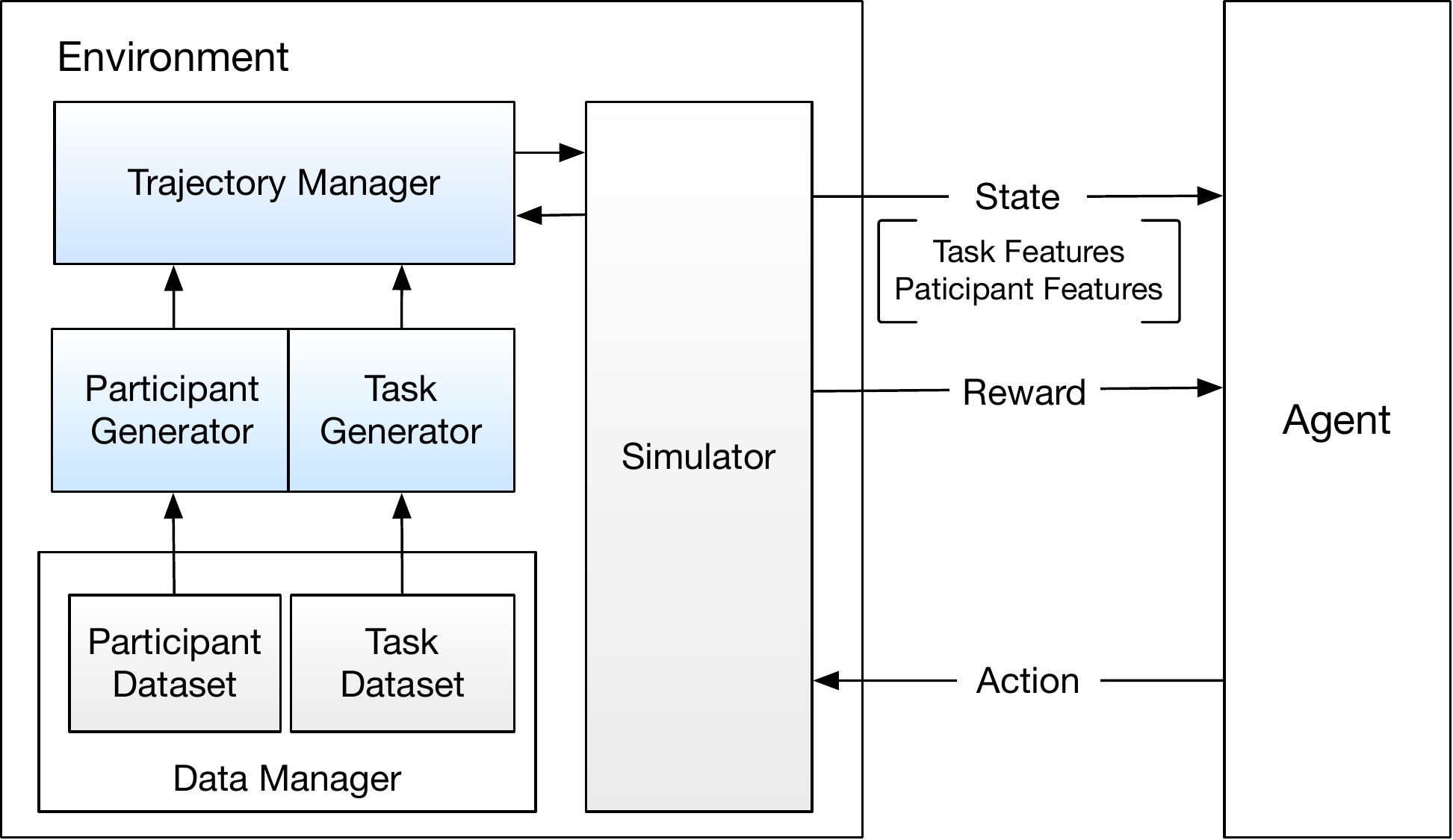}
    \caption{The framework of the simulated environment. 
    There are three modules in the simulated environment: the task and participant generation module, the simulator module, and the reward computing module.}
    \label{fig:DRLArc}
\end{figure}

In this section, we design a simulated environment for ride-sharing and show the framework in Figure \ref{fig:DRLArc}.

\subsubsection{Task and participant generation}
In our framework, the data manager provides the raw data of participants and tasks. The participant generator and the task generator preprocess the raw data and generate the participants and the tasks respectively by sampling from the data. The maximum limit on the number of the tasks and participants in each time interval can be customized according to the practical application scenario. Additionally, the trajectory manager maintains and evolves all the generated tasks and participants during the interaction with the agent. 

\subsubsection{Simulator}

The simulator extracts the task and participant features as the state of the environment to provide to the agent. Then the simulator receives and executes the actions of the agent. With the help of the trajectory manager, the simulator maintains and updates the states of all the participants and tasks synchronously. At last, the simulator computes and returns the reward to the agent.

In order to simulate a real ride-sharing environment for some complicated situations like traffic jams and different weather conditions, we set different moving speeds for the participants in different locations. In detail, the whole map is divided into a number of grids. We set the speeds in different grids on the map to simulate the traffic situation. For the weather variation, the moving speed of a participant in a grid changes along with time. In addition, a recruiter is possible to cancel the task after submitting this task in our environment.

\subsubsection{Reward computing}
The reward is a weighted sum of (1) \emph{distance cost} which is the distance between the participant and the recruiter when the task is assigned to the participant, (2) the total distance of all the completed tasks, (3) \emph{time cost} which is the total time cost of all the tasks, (4) the standard deviation of the \emph{distance cost} across different participants and the standard deviation of the \emph{time cost} across different tasks and (5) overall energy, which is the total energy consumed by all completed tasks. In this reward setting, we describe the overall profit by (1) and (2), the fairness by (3) and (4) and the effectiveness by (5). 

\begin{figure}[!t]
	\centering\includegraphics[width=3in]{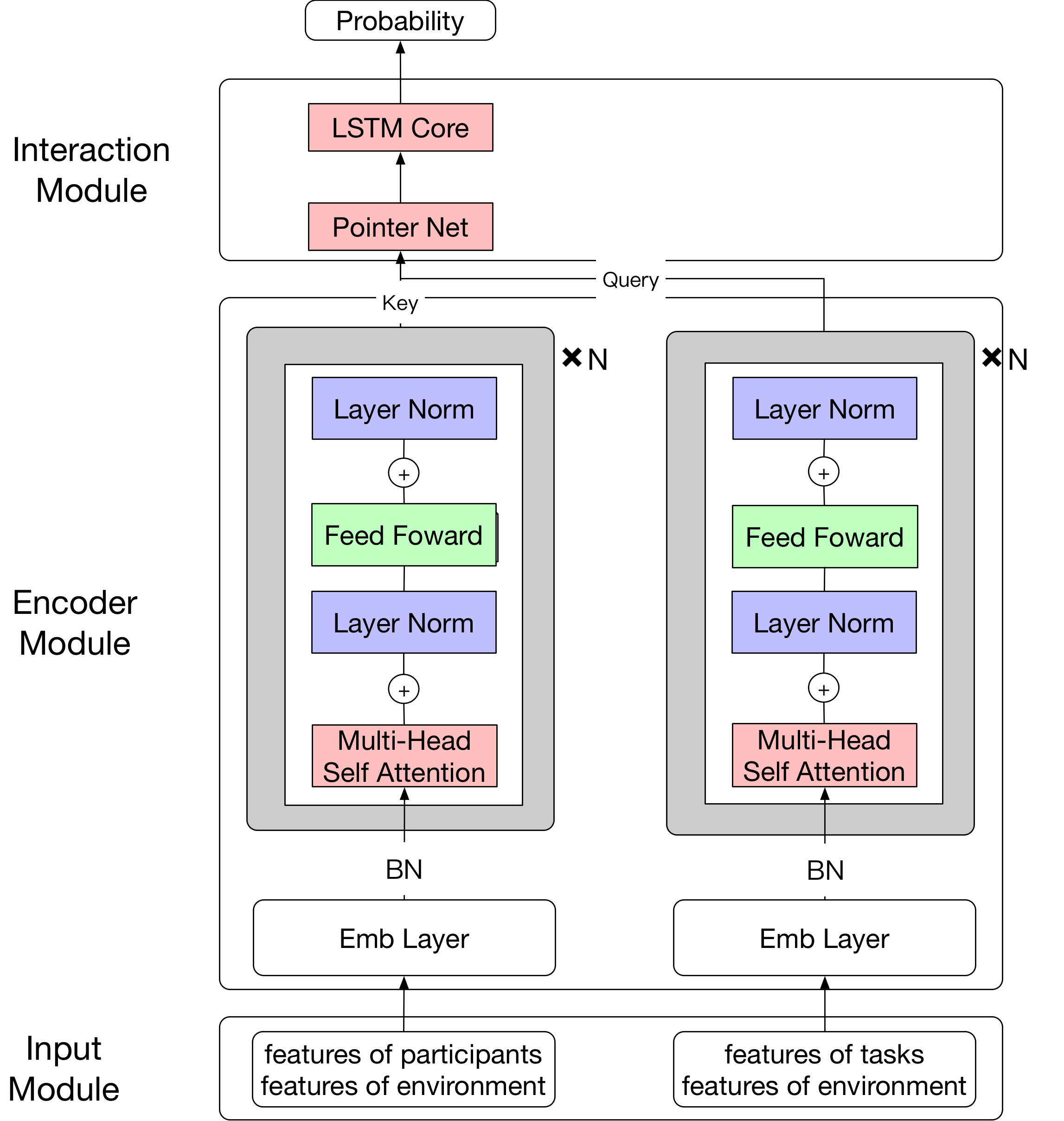}
	\caption{The architecture of the policy network. 
	There are three main modules in the architecture of the policy network: the input module, the task and participant encoder module, and the interaction module with a pointer network and an LSTM core. }
	\label{fig:MCS_RL_MA}
\end{figure}

\section{Policy Network}
ADRL is based on REINFORCE \cite{williams1992simple} with baselines, a classic policy gradient algorithm.
The policy in ADRL is approximated by a neural network whose architecture is specified in Figure \ref{fig:MCS_RL_MA}. The architecture consists of four modules: the input module, the task and participant encoder module, the interaction module with pointer network and LSTM core, and the greedy selection module. 
\subsubsection*{Input module}
In the input module, the task, participant, and environment are extracted as corresponding features. The features of the $i$-th task are denoted as $\bm{t}_i$. The features of the $j$-th participants are denoted as $\bm{p}_j$.
The environment features are included in the participant and task features. The features are transformed to task and participant embeddings by the embedding layers.
To distinguish between different participants and tasks, batch normalization (BN) layers \cite{ioffe2015batch} are appended to the embedding layers. In the previous work \cite{santurkar2018does}, the batch normalization layer is used to address the internal covariate shift problem. In our model, it helps to distinguish among the participant (or task) embeddings. We found that these BN layers make the training process more robust to weight initialization and different learning rates.

\subsubsection*{Task and participant encoder module}

The encoder of the task or the participant is composed of multiple identical blocks. Each block is composed of four layers: a multi-head self-attention layer \cite{Vaswani2017Attention}, a fully connected feed-forward layer, and two layer norm layers \cite{ba2016layer}. 

There are two types of self-attention layers: the participant self-attention layers and the task self-attention layers. In the former type of layers, the queries, keys and values are all participant embeddings. The pairwise interaction between two participants in $\{\bm{p}_i\}_{i=1:n}$ is represented by $\bm{p}_i \cdot \bm{p}_j$. Then the participant embedding $\bm{p} := \{\bm{p}_i\}_{i=1:n}$ is updated by accumulating information from all the interactions, $\bm{p} \leftarrow \text{softmax}(\bm{p}\cdot \bm{p}^T) \bm{p}$ in the encoder module. The latter type of layers process in a similar manner. The activation functions are ReLU functions \cite{nair2010rectified}.

\subsubsection*{Interaction module with pointer network and LSTM core}

The pointer network and the LSTM core are adopted to select suitable participants for each task.
Using the pointer network, each task can interact with every participant and this benefits the combinational optimization. In the pointer network, the interaction between the $j$-th participant and the $i$-th task is denoted as
\begin{equation}
u_j^i = \bm{p}_j\cdot \bm{t}_i.
\end{equation}

Then, we use the LSTM core to select suitable participants for each task one by one, following an order according to the arrival time. For the $i$-th task, the LSTM core gives the probability that we select each participant in $\bm{P}$.
\begin{equation}
\mathcal{P}(\bm{P}; t_i) = \text{softmax}( LSTM(\bm{u}^i|\bm{u}^1, \cdots, \bm{u}^{i-1})),
\end{equation}
where $\bm{u}^i$ is concatenation of $\{u_j^i\}_{j=1}^n$. Each element in $\mathcal{P}(\bm{P};t_i) \in \Delta(\bm{P})$ denotes the probability that the corresponding participant in $\bm{P}$ is selected by the $i$-th task.
By considering the participants selected by preceding tasks, the LSTM core alleviates the \emph{collisions} between the participant selections on different tasks. Besides, the LSTM core makes the training process more stable, which is observed from our experiments. These innovations in the policy network improve the quality of the representations learned, leading to significant performance gains.

\subsubsection*{Greedy selection module}
In the $h$-th time interval, the previous module outputs the probability $\mathcal{P}(\bm{P}_h;t_i)$ that each participant is selected to accomplish the task $t_i$. In the greedy selection module, we sequentially process the tasks in the same aforementioned order. 
For each task $t_i \in \bm{T}_h$, we select suitable participants in $\bm{P}_h$ with the probability $\mathcal{P}({P}_h;t_i)$, and then remove the selected participants from $\bm{P}_h$.

\section{Auxiliary-Task Training Process}

\begin{figure}[!t]
    \centering\includegraphics[width=3.2in]{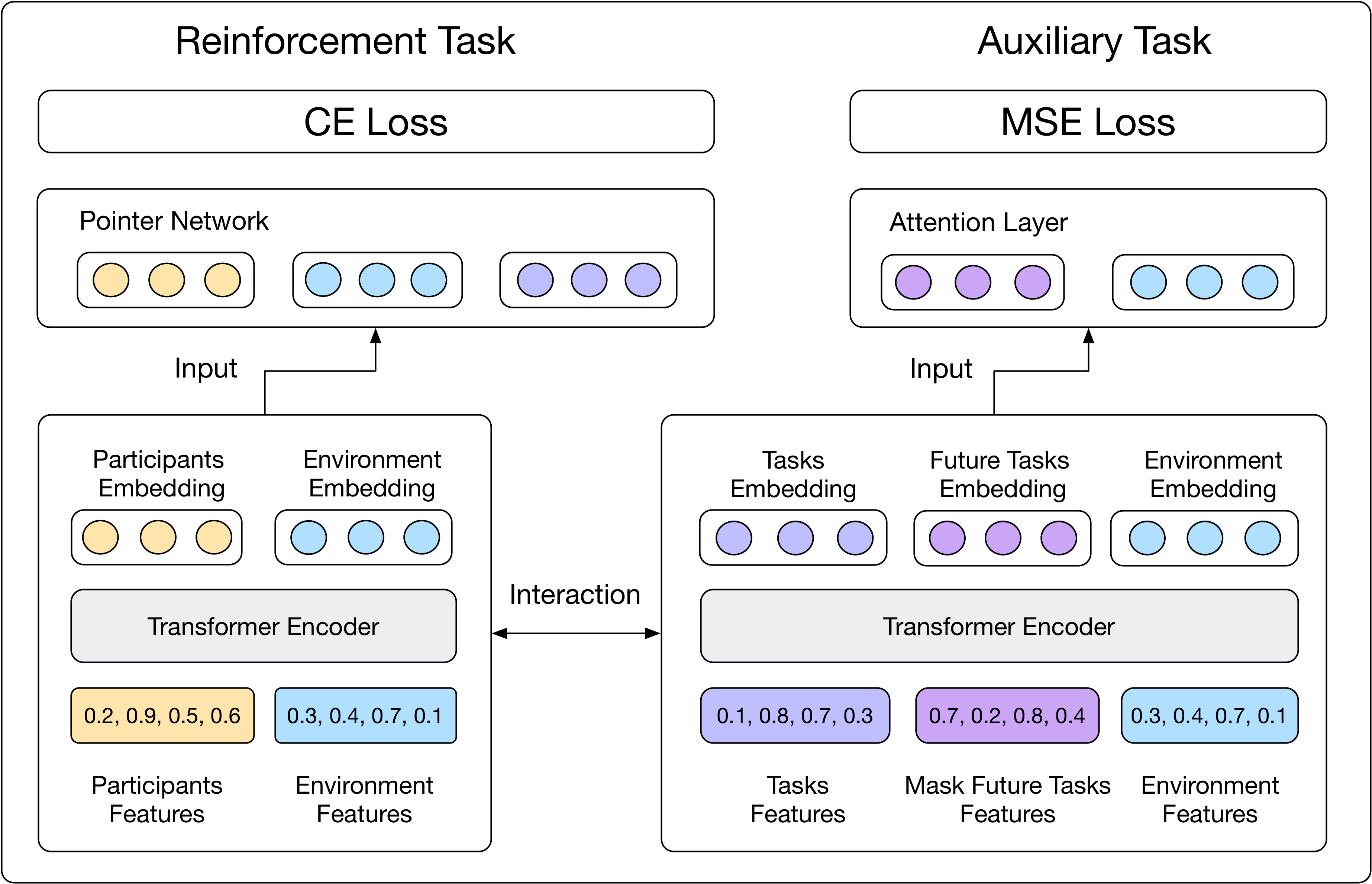}
    \caption{The auxiliary-task training process. In this process, we train the network with an additional loss which is the mean squared loss between the predicted tasks and the truth.}
    \label{fig:MT_Learn}
\end{figure}

Conventional model-free reinforcement learning suffers from sample inefficiency. To accelerate the training process and improve the final performance, we implement an auxiliary-task training process which is shown in Figure \ref{fig:MT_Learn}.

Consider a case where there are no adequate participants around the location of a recruiter. The system has two choices: 1) accept this task and allocate a participant from a far area to accomplish the task, or 2) do not accept this task. However, this decision depends on whether there will be an increasing number of tasks appearing in this area in the future. Accordingly, it is easier to make a good decision when the future task appearance is predicted.

Based on the above analysis, the model should not only take account of the past tasks but also predict future tasks in the decision process. The prediction of future tasks provides more information for making decisions. 

The model predicts the future tasks by an additional attention layer appended to the task encoder module, where the task embeddings from the task encoder are decoded as the embeddings of the predicted tasks.
Moreover, we train the network with an additional loss which is the mean squared loss between the predicted tasks and the truth (We use $N(\theta)$ to denote the loss of the auxiliary task associated with the policy $\pi_\theta$.). As a consequence, the model selects better participants for the current tasks through the prediction of future tasks.

The objective of reinforcement learning is to maximize the cumulative reward
\begin{equation}
J(\theta) = E_{\tau \sim p_{\theta}(\tau)}[r(\tau)],
\end{equation} 
where $\theta$ is the parameter of policy network, $\tau$ is a sequence of $s_0, a_0, s_1, a_1, \cdot \cdot \cdot$, $p_\theta(\tau)$ is the probability that the sequence $\tau$ is sampled when we adopt the policy $\pi_{\theta}$, and $r(\tau)$ is cumulative reward of the trajectory.

 Therefore, the loss of the auxiliary-task training process is to minimize
\begin{equation}
L(\theta) = -J(\theta) + \alpha \cdot N(\theta),
\end{equation}
where $\alpha$ is a hyper parameter.

\section{Experiments}
\begin{table}[!t]
	\centering
	\caption{The statistics of the trajectory dataset and the hyperparameters used in the experiments.}
	\begin{tabular}{p{0.6\columnwidth}<{\centering}|p{0.3\columnwidth}<{\centering}}
		\toprule
		Number of participants & 564,516 \\
		Number of trajectories & 26,206 \\
		Number of points & 616,928 \\
		Number of passengers &  41,989 \\
		Total distance & 66,546.73 km \\
		Total duration &  21,602 hours \\
		Total fare amount &  3,249,543 dollars \\
		Total amount &  400,689.55 dollars \\
		Number of transformer layers & 8\\ 
		$\alpha$ in equations (6) & $0.1$ \\
		\hline
	\end{tabular}
	\label{table:dataset}
\end{table} 
\begin{table}[!t]
    \centering
        \caption{The ablation study in the atom task. 
    In Model with SL, we perform a supervised learning over the optimal solutions using the policy network architecture in Figure \ref{fig:MCS_RL_MA}.
    In Model with RL, we perform the reinforcement learning with the same policy network.
    Then, we respectively drop the BN layer,
     (Model without BN Layer)
    the LSTM layer 
     (Model without LSTM Layer) 
    and the dropout layer;
     (Model without Dropout Layer);
    we replace the transformers with MLPs 
      (Model with MLP Only) 
    or use only one layer in the transformers.
     (Model with One Layer Transformer).
    }
    \begin{tabular}{m{120pt}<{\centering} m{50pt}<{\centering}}
        \toprule
        Solution & Reward \\
        \midrule
        Optimal Solution & 0.088\\
        Model with SL & 0.087 \\ 
        Model with RL & 0.086 \\ 
        Model without BN Layer &  0.070\\
        Model without LSTM Layer & 0.077\\
        Model without Dropout Layer & 0.079\\
        Model with One Layer Transformer & 0.077\\
        Model with MLP Only & 0.066\\
        \hline
    \end{tabular}
    \label{table:atom}
\end{table}  
The performance of ADRL is evaluated by a series of experiments in the simulated environment. We first introduce the experiment setups and some baseline methods in detail. Then, we present and compare the evaluation results of our algorithm to the baseline methods and the ablated variants.

\begin{table*}[tp]
    \centering
    \caption{The performance comparison between the baselines and ADRL in fairness-first reward settings.
    In the fairness-first reward setting, the weight of the fairness reward component is equal to the sum of the weights of the other components. In the environment setting, $s$ represents the number of steps (time intervals), $t$ represents the number of tasks and $p$ represents the number of participants.}
        \begin{tabular}{ccccccccc}
            \toprule
            \multicolumn{3}{c}{Environment Settings}&\multicolumn{6}{c}{Fairness-First Reward}\cr
            \cmidrule(lr){1-3} \cmidrule(lr){4-9}
            $s$&$t$&$p$&NPF&NAPF&WPF&RWM&PPOWM&ADRL\cr
            \midrule
2 &2 &10 &\textbf{3.828}  &\textbf{3.828}  &3.099  &2.515  &2.605  &3.790 \cr
2 &5 &10 &5.247  &5.298  &5.335  &5.642  &5.341  &\textbf{7.576} \cr 
5 &5 &5 &1.329  &1.993  &2.898  &4.331  &5.287  &\textbf{6.368} \cr 
5 &5 &15 &10.962  &12.425  &8.673  &15.269  &13.281  &\textbf{16.832} \cr 
10 &5 &10 &17.096  &22.029  &21.407  &22.510  &21.926  &\textbf{23.186} \cr 
5 &10 &15 &25.473  &20.310  &18.586  &24.310  &20.006  &\textbf{27.172} \cr 
10 &10 &20 &28.996  &37.868  &41.356  &42.747  &36.171  &\textbf{47.832 } \cr 
20 &5 &30 &9.117  &14.299  &51.235  &54.834  &37.885  &\textbf{61.852} \cr 
            \bottomrule
        \end{tabular}
    \label{table:performance_comparison_1}
\end{table*}

\begin{table*}[tp]
    \centering
    \caption{The performance comparison between the baselines and ADRL in energy-first reward setting. In the energy-first reward setting, the weight of the energy efficiency reward component is equal to the sum of the weights of the other components. }
        \begin{tabular}{ccccccccc}
            \toprule
            \multicolumn{3}{c}{Environment Settings}&\multicolumn{6}{c}{Energy-First Reward}\cr
            \cmidrule(lr){1-3} \cmidrule(lr){4-9}
            $s$&$t$&$p$&NPF&NAPF&WPF&RWM&PPOWM&ADRL\cr
            \midrule
2 &2 &10 &3.019  &3.019  &2.371  &2.626  &2.295  &\textbf{3.087} \cr
2 &5 &10 &4.600  &4.636  &4.831  &5.944  &4.917  &\textbf{7.353}\cr
5 &5 &5 &2.824  &2.684  &4.241  &5.397  &5.936  &\textbf{6.387} \cr
5 &5 &15 &10.453  &11.406  &9.030  &15.657  &12.752  &\textbf{19.304} \cr
10 &5 &10 &17.046  &20.254  &20.279  &24.919  &20.908  &\textbf{25.995} \cr
5 &10 &15 &25.092  &24.430  &22.130  &26.467  &20.724  &\textbf{30.817} \cr
10 &10 &20 &32.309  &39.970  &40.399  &\textbf{48.445}  &38.075  &48.075 \cr
20 &5 &30 &19.504  &21.849  &48.063  &55.117  &37.007  &\textbf{57.683} \cr
            \bottomrule
        \end{tabular}
    \label{table:performance_comparison_2}
\end{table*}

\begin{table*}[tp]
    \centering
    \caption{The performance comparison between the baselines and ADRL in profit-first reward setting. In the profit-first reward setting, the weight of the profit reward component is equal to the sum of the weights of the other components.}
        \begin{tabular}{ccccccccc}
            \toprule
            \multicolumn{3}{c}{Environment Settings}&\multicolumn{6}{c}{Profit-First Reward}\cr
            \cmidrule(lr){1-3} \cmidrule(lr){4-9}
            $s$&$t$&$p$&NPF&NAPF&WPF&RWM&PPOWM&ADRL\cr
            \midrule
2 &2 &10 &2.372 &2.372 &1.796 &\textbf{2.682} &2.649 &2.513\cr
2 &5 &10 &4.083 &4.105 &4.458 &6.154 &5.206 &\textbf{7.269}\cr
5 &5 &5 &4.020 &3.237 &5.160 &6.597 &7.028 &\textbf{7.996}\cr
5 &5 &15 &6.246 &7.790 &8.757 &16.486 &12.879 &\textbf{21.934}\cr
10 &5 &10 &17.006 &18.833 &19.532 &27.120 &23.085 
 &\textbf{28.115}\cr
5 &10 &15 &25.261 &22.608 &20.606 &29.037 &22.990 &\textbf{34.856}\cr
10 &10 &20 &34.959 &41.651 &39.935 &49.786 &41.796 &\textbf{51.899}\cr
20 &5 &30 &27.814 &27.889 &41.520 &55.096 &41.893 &\textbf{60.694}\cr
            \bottomrule
        \end{tabular}
    \label{table:performance_comparison_3}
\end{table*}
\subsection{Datasets and Experiment Setups}

The datasets used in the stimulated experiment are the NYC Taxi dataset and the Uber Trips dataset \cite{Todd2018data}, which are both open-sourced \cite{Todd2018git}. Table \ref{table:dataset} lists the scale of the whole dataset.
It was collected by publicly available taxi and Uber trajectories in NYC from 2009 to 2018, covering billions of individual trips. In the taxi dataset, each trip of a vehicle contains not only the coordinates and timestamps of the pickup and drop-off locations but also the trip distances and the detailed fares. The data of Uber Trips is different from that of NYC Taxi. The coordinates of a Uber vehicle are given in the order of time, and thus the one-day trips are recorded with discrete timestamps. The pickup and drop-off coordinates are not recorded. These two sources of data are preprocessed and then the tasks and participants are sampled from the data in the simulated environment.

In addition, we use an eight-layer transformer in the policy network. We set $\alpha$ to $0.1$, the learning rate to $0.1$ and the batch size to $640$. 

\subsection{Baseline Methods}
In this section, we introduce a series of baseline methods that are adopted from \cite{Pu2016Crowdlet} and \cite{Suhr:2019}.

\noindent\textbf{Nearest participant first (NPF).} NPF is an algorithm that allocates a pending task to the nearest participant, which is a simple but effective participant selection method for ride-sharing tasks. 

\noindent\textbf{Nearest available participant first (NAPF).} NAPF is an improved version based on NPF \cite{Pu2016Crowdlet}. The participants allocated to a time-consuming task may be unavailable in a long period of time. To avoid selecting the participants that are unavailable in a long time to complete the task, NAPF selects the nearest and currently available participant to complete the task.

\noindent\textbf{Worst-off participant first (WPF).} Given a task, the WPF algorithm selects the participants allocated with the fewest tasks within a given distance from the task. This algorithm pays attention to both the overall profit and the fairness of the system. 

\noindent\textbf{Reinforce algorithm with multi-layer perceptron (we denote it as MLP) (RWM).} The RWM method uses the Reinforce algorithm in which we adopt MLP  as the policy network. 

\noindent\textbf{Proximal policy optimization algorithm with MLP (PPOWM).} The PPOWM method uses the Proximal policy algorithm \cite{schulman2017proximal} in which we adopt MLP as the policy network.   

\noindent\textbf{Reinforce algorithm with the same policy network in ADRL (DRL).} The DRL method uses reinforce algorithm in which the policy network is the same as ADRL. 

\subsection{Atom Task and Ablation Study}
We first introduce a minimal ride-sharing task, called atom task. In this task, there is one ride-sharing task with two participants in a particular time interval, whose optimal solution is easy to gain. We conduct two experiments on the atom task.

1) We first do a supervised learning (SL) over the optimal solution, which is used to test the model capacity. Then we compare the performance of SL and RL. We show the result in Table \ref{table:atom}. We find that the performance of the RL method is comparable to that of the SL method, which indicates that the RL model is capable to represent a near-optimal solution for the atom task.  

2) Then, we do an ablation study to evaluate the effectiveness of the policy network in ADRL. As shown in Table \ref{table:atom}, we test the effectiveness of the designs in our deep learning architecture (such as Transformer, BN, LSTM, and the dropout layer). We first replace the Transformer with MLP and one layer Transformer in the policy network. Then, the performance declines, which demonstrates that the eight-layer transformer is important in the policy network. In addition, we drop the BN, LSTM, and dropout layer in our deep learning architecture respectively. The result demonstrates that the ablation of any design in our model results in performance degradation.

\begin{figure*}[tp]
    \centering\includegraphics[width=6.5in]{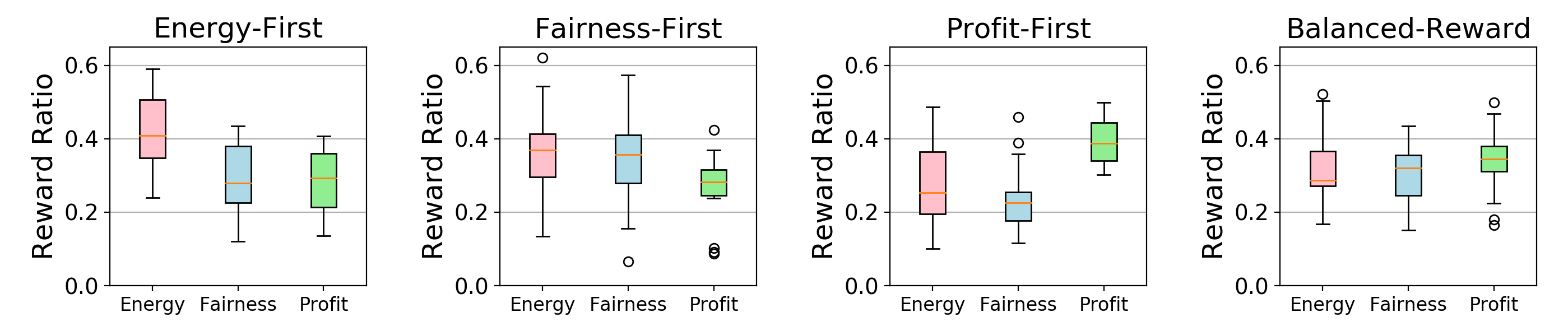}
    \caption{Reward proportion analysis. We analyze the reward proportion of three goals attained by the policies trained by ADRL in three reward settings respectively. Additionally, we do the experiments in the reward setting that the weight of each goal is equal. The statistics of these experiments are shown in the sub-graph \emph{Balanced-Reward}.}
    \label{fig:Reward_cmp}
\end{figure*}

\subsection{The Performance Evaluation} 

We compare ADRL with the baseline methods under three types of rewards and eight different environment settings, and show the performances in Table \ref{table:performance_comparison_1}, \ref{table:performance_comparison_2} and \ref{table:performance_comparison_3}. 
For the three types of rewards, we set different weights for the reward components to emphasize the fairness, the overall profit or the energy efficiency respectively. For instance, we set the weight of the component that describes the fairness equal to the sum of the weights of other components in the fairness-first reward setting. This reward setting pays more attention to the fairness. The result shows that ADRL attains a comparable or better policy than the baseline methods across different settings, which indicates that ADRL is effective for different goals. In addition, we conduct experiments with eight different environment settings. These settings represent different levels of complexity in the MCS systems. In some of the MCS systems, there are adequate participants to complete with each other. In some of the MCS systems, the number of steps is large which requires a more careful and complex planning.

In these experiments, ADRL is better than NPF, NAPF, WPF, RWM and PPOWM in 21 out of 24 settings. In three of the settings 
(the number of time steps, tasks and participants are 2, 2, 10 in the fairness-first and profit-first settings and 10, 10, 20 in the energy-first setting), ADRL is worse than some of the baselines, due to the fact that the baseline methods can attain a near-optimal solution in these simple situations (as in the atom task) or the reward setting is not intricate. In the experiments, we observe that in easier tasks where there are adequate number of participants, ADRL is only slightly better than the baselines. Otherwise, ADRL has a significantly better performance than the baselines. This demonstrates that ADRL is better to deal with complex combinational optimizations in the participant selection problem.

\begin{figure*}[tp]
    \centering\includegraphics[width=6.5in]{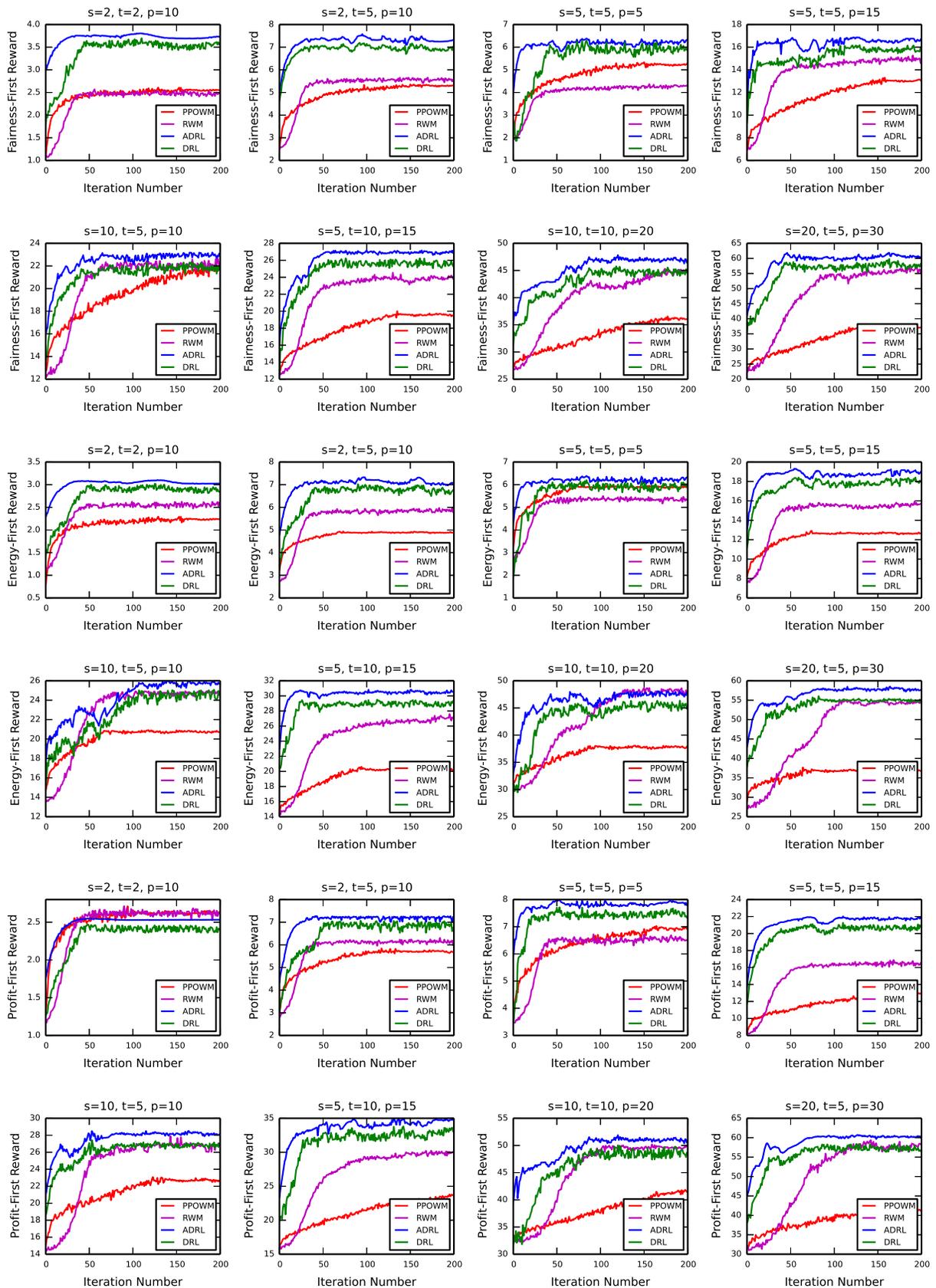}
    \caption{The performance of ADRL (ours), DRL, RWN and PPOWN}
    \label{fig:Iteration}
\end{figure*}

In addition, we compare ADRL with RWM and PPOWM in Figure \ref{fig:Iteration}. The performance of ADRL is better than the RWN and PPOWM algorithms. Furthermore, instead of RWN and PPOWN, the ADRL can attain a good performance by a few iterations. It indicates that ADRL needs fewer iterations to train policy network than baselines.

In Figure \ref{fig:Reward_cmp}, we present the reward proportions of the three sub-goals attained by the policies trained by ADRL in the fairness-first, energy-first and profit-first reward settings. 
Additionally, we conduct the experiments in the reward setting that the weights of these sub-goals are equal. 
The statistics of this set of experiments are shown in the sub-graph \emph{Balanced-Reward}. 
In these different reward settings, the policies pay attention to different sub-goals. 
For instance, in the fairness-first setting, the reward component corresponding to the fairness from the trained agent takes up a large proportion of the total reward, which demonstrates that the policy is indeed a fairness-first policy.

\subsection{Auxiliary-Task Training Process}  
In this comparative experiment, we test the effectiveness of the auxiliary-task in ADRL. We show the result in Figure \ref{fig:Iteration}. The ADRL agent is trained to optimize the policy and predict the future tasks. The DRL agent is only trained to optimize the policy. The curves are the averaged cumulative reward on each epoch during the training. In the entire training process, the performance of ADRL is better than DRL. Furthermore, the ADRL agent can obtain a good performance in the begining of the training process and obtain a more stable performance than the DRL agent in the end of training process. 
The experiment discloses that the auxiliary-task training process accelerates the training of the DRL and achieves a better performance. 

\section{Related Work}
\subsubsection*{Mobile crowdsourcing and participant selection} Previously, several approximate algorithms based on specific assumptions have been proposed to solve the MCS problem. 
However, these assumptions are relatively strong. 
For example, \citeauthor{Zhang2014CrowdRecruiter} \cite{Zhang2014CrowdRecruiter} assumes that the income is equally paid to each participant; \citeauthor{Pu2016Crowdlet} \cite{Pu2016Crowdlet} requires the information of each participant. 
Moreover, estimating the reliability of participants is also significant in MCS \cite{Bonald2017AMO}. 

In the specific MCS task, ride-sharing task, \citeauthor{Zhang:2017KDD} \cite{Zhang:2017KDD} propose a novel system to dispatch taxis to finish multiple tasks aiming to optimizes the overall travel efficiency. 
In \cite{Xu:2018:kdd}, an order dispatch problem is modeled as sequential decision-making and a Markov decision process method is deployed. In \cite{lin2018efficient}, a multi-agent reinforcement learning network is used to tackle the large-scale fleet management problem, which improved the system satisfaction and increased the utilization of transportation resource \cite{guyen2017collective}. The travel time prediction model is presented in \cite{zhengW2018kdd}. 
Moreover, \citeauthor{qin2019deep} \cite{qin2019deep} present a simulation-based human-computer interaction of deep reinforcement learning in action on order dispatching and driver re-positioning for ride-sharing. They use deep reinforcement learning to train agents (drivers) to cooperate to achieve higher objective values.
As many researchers focus on the ride-sharing task, there is adequate data to supply a simulated environment and different baselines to compare with. Moreover, we can optimize for multiple goals in the ride-sharing tasks, such as profit, fairness and energy-efficient. Therefore, we test ADRL on the ride-sharing task in our paper. However, our method can be easily applied in other MCS tasks. 

\subsubsection*{Deep reinforcement learning with multi-head attention and Pointer Networks} 
Another relevant research topic is deep reinforcement learning with multi-head attention and Pointer Net. 
In \cite{Zambaldi2019Relational}, a multi-head attention based RL algorithm is proved to perform relational reasoning over structured representations successfully. 
In \cite{vinyals2015pointer}, Pointer Net is trained to output satisfactory solutions to three combinatorial optimization problems. 
In AlphaStar \cite{vinyals2017starcraft}, the agent trained via multi-agent reinforcement learning with multi-head attention and Pointer Net beats 10 professional human players. 
It motivates us to use these advanced techniques to exploit the inherent structures in the MCS environment and learn better representations.
Similar to \cite{vinyals2015pointer}, we use Pointer Net to perform the combinatorial optimization problem, i.e., selecting suitable participant to complete the tasks. 

\subsubsection*{Auxiliary-task learning in machine learning} 
Learning auxiliary tasks simultaneously along with the main task can help learn a representation with better generalization.
It leverages the domain knowledge from other tasks (e.g., self-supervised learning or unsupervised learning) to improve the performance of the main task.
For example, in natural language processing, language models \cite{radford2019language,brown2020language} which are train without explicit supervision can achieve good performance in many tasks in the zero-shot or few-shot settings.
In reinforcement learning, the auxiliary-task is also successful. 
In the model-based settings, world models can be trained to learn a compact representation of the environment.
\citeauthor{ha2018world} \cite{ha2018world} adopt a two-stage (or iterative) process where a world model is first learned in a unsupervised manner from samples and then used to train the policy.
However, learning the auxiliary tasks simultaneously (as we do) often leads to a better performance \cite{veeriah2019discovery}, such as Dreamer \cite{hafner2019dream} and PlaNet \cite{hafner2019learning}. 
In the model-free settings, many algorithms construct various auxiliary tasks to improve performance, such as predicting the future states \cite{guo2020bootstrap,jaderberg2016reinforcement}, predicting various value functions with different cumulants \cite{veeriah2019discovery} or policies \cite{dabney2020value}, or recognizing transformations to the image-based states \cite{srinivas2020curl}.
In MCS, a multi-view network is applied in taxi demand prediction \cite{Yao2018DeepMS} and a multi-task representation learning model is proposed for travel time estimation \cite{Li2018kdd}. Similarly, we use the auxiliary task in our method to facilitate the representation learning.

% \begin{figure}[!t]
%     \centering\includegraphics[width=2.1in]{MT_Process.pdf}
%     \caption{Comparison of the training process with and without auxiliary-task training.}
%     \label{fig:MT_Process}
% \end{figure}

\section{Conclusion}

In conclusion, we formulate the participant selection problem as a reinforcement learning problem and propose the auxiliary-task deep reinforcement learning method (ADRL) to solve it, which is shown to be a good training method to achieve flexible and compound goals in MCS. Besides, we release a simulated MCS environment based on real datasets. To facilitate further research and reproducibility of our results, we base the simulated environment on the OpenAI Gym interface and publish it online. At last, we conduct extensive performance evaluations and the experimental results demonstrate that ADRL achieves better performance and accelerates the training process of deep reinforcement learning.

There are two future directions based on our work. 
First, it is worthy to investigate the detailed reason why the auxiliary-task training process improves the sample efficiency in RL. 
Second, we could introduce more realistic factors such as the road conditions and the user profiles into the simulated environment to model a more complex world.

\section{Acknowledgement}
This work is supported in part by the National Natural Science Foundation of China, under Grant No. 61672276, the Key Research and Development Project of Jiangsu Province under Grant No. BE2019104, and the 
Collaborative Innovation Center of Novel Software Technology and Industrialization, Nanjing University.

\bibliographystyle{ACM-Reference-Format}
\balance
\bibliography{adrl-ref}

\end{document}